\documentclass[letterpaper, 8 pt, conference]{ieeeconf}  

\IEEEoverridecommandlockouts                              
\usepackage{caption}
\usepackage{esvect}
\usepackage{graphicx}
\usepackage{subfig}
\usepackage{mathptmx} 
\usepackage{times} 
\usepackage{amssymb}
\usepackage{mathtools}

\usepackage{bm}
\usepackage[bookmarks=false]{hyperref}
\hypersetup{
    colorlinks=true,
    linkcolor=blue,
    filecolor=magenta,      
    urlcolor=black,
}
\usepackage{ragged2e}
\usepackage{listings}
\newcommand{\dollar}{\mbox{\textdollar}}
\usepackage{textcomp}

\title{\LARGE \bf
A Real-time Control Approach for Unmanned Aerial Vehicles using Brain-computer Interface*
}

\author{Ravi M. Vishwanath$^{1}$, Saumya Kumaar$^{1}$ and 
S N Omkar$^{1}$
\thanks{$^{1}$ All authors are with the Indian Institute of Science, Bangalore}
\thanks{* This project was funded by the JATP, Defense Development Reseach Organization (DRDO), Govt. of India}
\thanks{Copyrights - 978-1-5386-5323-4/18/$\dollar$31.00 \textcopyright 2018 IEEE}
}

\begin{document}

\maketitle

\thispagestyle{empty}
\pagestyle{empty}

\begin{abstract}
Brain-computer interfacing (BCI) is a technology that is almost four decades old and it was developed solely for the purpose of developing and enhancing the impact of neuroprosthetics. However, in the recent times, with the commercialization of non-invasive electroencephalogram (EEG) headsets, the technology has seen a wide variety of applications like home automation, wheelchair control, vehicle steering etc. One of the latest developed applications is the mind-controlled quadrotor unmanned aerial vehicle. These applications, however, do not require a very high-speed response and give satisfactory results when standard classification methods like Support Vector Machine (SVM) and Multi-Layer Perceptron (MLPC). Issues are faced when there is a requirement for high-speed control in the case of fixed-wing unmanned aerial vehicles where such methods are rendered unreliable due to the low speed of classification. Such an application requires the system to classify data at high speeds in order to retain the controllability of the vehicle. This paper proposes a novel method of classification which uses a combination of Common Spatial Paradigm and Linear Discriminant Analysis that provides an improved classification accuracy in real time. A non-linear SVM based classification technique has also been discussed. Further, this paper discusses the implementation of the proposed method on a fixed-wing and VTOL unmanned aerial vehicles. 
\end{abstract}
\section{INTRODUCTION}

Research in neurological studies hit a roadblock when the need arose to understand brain waves. The complexity of such waves could only be studied using advanced computational tools. Brain-Computer Interfaces (BCIs) or Mind-Machine Interfaces (MMI) were born out of the need to capture and analyze the signals on computers. BCI is an artificial system that incorporates the communication between the brain and an external device or a computer. [1] [2] In this system the activity of the brain during a certain action is tapped into using electroencephalogram (EEG) devices which are sent to central system to be processed
and extract them into control signals, initially devised for rehabilitation to help people regain motor skills that are lost or absent now promises a new field of research for both medical and Engineering applications. The first research on BCIs took place in 1970 at University of California at Los Angeles which was aimed at evaluating the ability of BCI to bolster neuroprosthetics [3][4].

BCI gained further traction when it was used to treat locked-in syndromes and neuromuscular disconnections. Auditory, visual and facial muscles often lose the reliability and are often exhausted when in frequent or prolonged use [1]. Every thought or action gives rise to a particular electrical activity. A state of daydreaming or deep meditation emits delta waves (0-3 Hz). Sleeping emits theta waves (3-7 Hz). A state of consciousness emits alpha waves (8-12 Hz). Engagement in a particular activity or problem-solving emits beta waves (12-38 Hz) [5][12]. An EEG uses this frequency domain feature to display the users’ intent in terms of brain waves onto an external device to monitor and control an external device. The very first of BCI systems were the P300 spellers which could read the brain signals and read letters for communication between the patients who were physically challenged [4]. California medical center came up with brain-controlled prosthetic legs with the idea of helping the patients with spinal cord injuries walk. The system could be used by the patient to move the prosthetic in real time [5]. A similar system was developed in Washington to restore the hands for the people with neural disconnection [6]. A BCI-controlled wheelchair was developed for the patients who could not use the joystick or the same. Different models of the wheelchair were developed depending on the amount of control left to the user’s discretion. [7] The scope of research of BCI was initially confined to only medical applications for detection of different brain states such as alertness, emotion, attention; however, it has now extended to include engineering and industrial applications as well. In a paper proposed by Bastian Venthur et. al, the concentration level of an operator in a factory was examined over a period of time. They prepared a model that would access the response time of an operator to the different levels of alert messages. This would serve helpful in avoiding accidents that could otherwise occur due to low alertness or fatigue [8]. The introduction of BCI’s in the field of home automation has been explored by Wei Tuck Lee, et al, wherein they have prepared a virtual home environment in which an individual can control appliances directly with the mind [9]. Brain-Computer Interface is also found promoting itself into the field of Aerospace Engineering, there have been successful attempts made to control a multi-rotor unmanned aerial vehicle (UAV) using BCI. A successful prototype has been developed wherein a quadcopter is made to maneuverer upward, downward, right and left directions depending on the command received by the pilot’s brain signals [10]. In another successful attempt by Vijayendra \textit{et. al.} [18], the authors have demonstrated 95\% accuracy in brain-computer interface based control of UAVs, with the trade-off between accuracy and real-time implementation. The proposed approach is very efficient but has limited rate of 25 Hz. Moreover, there has been no attempt so far to control a fixed-wing UAV due to the complexity of streaming of data at a very high rate and their subsequent classification of the EEG signals that is requisite for high-speed UAV control. This paper aims to solve the complexities involved in the implementation of a BCI controlled fixed wing aircraft. The main contribution of this paper is designing the control methodology for maneuvering a delta-winged UAV using Common Spatial Paradigm (CSP) and Linear Discriminant Analysis (LDA) of the EEG signals and its implementation. These methods, with data processing rate of around 97 Hertz, provide a rapid classification platform for establishing a stable control link to the UAV and also achieve a high level of classification accuracy of 85\%.

\section{METHODOLOGY}
\subsection{Experimental Subjects}

There were a total of 14 subjects involved in this experimental study, out of which 5 were female and 9 male subjects, all aged between 18-24 years. The Selected Subjects did not have any prior experience in any BCI/HMI related tasks and a written consent of their participation in the study was submitted to the ethical committee of Indian Institute of Science.

\subsection{Acquisition Protocol}

A protocol was defined and followed for acquisition of EEG data from the subjects. 
The protocol dictates four motor imagery task be performed with rest breaks incorporated in 
between each tasks. This ensures that the motor imagery tasks are reinforced and the retention of information in the signal is sufficient. Fig.1 visualizes the protocol followed for data acquisition. 

\begin{figure}[h]
\centering
\includegraphics[width=3.4 in]{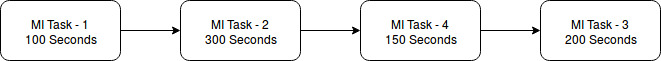}
\caption{Workflow of MI tasks}
\label{fig_chi_dot}
\end{figure}

The Motor Imagery (MI) Acquisition procedure is split into four tasks. Task 1 involves visualizing left-hand motion without any physical movement. Task 2 is the same as Task 1 but is performed with the right-hand. Task 3 requires the subject to visualize left-hand movement along with the movement of fingers and elbow. Task 4 replicates Task 3 but is performed with the right-hand. The EEG data acquired during these 4 tasks is depicted in Fig. 2. The variations in the EEG data are visually discernable and the accuracy with which the features can be extracted is solely depended on the algorithm used.

\begin{figure}[h]
\centering
\includegraphics[width=3.4 in]{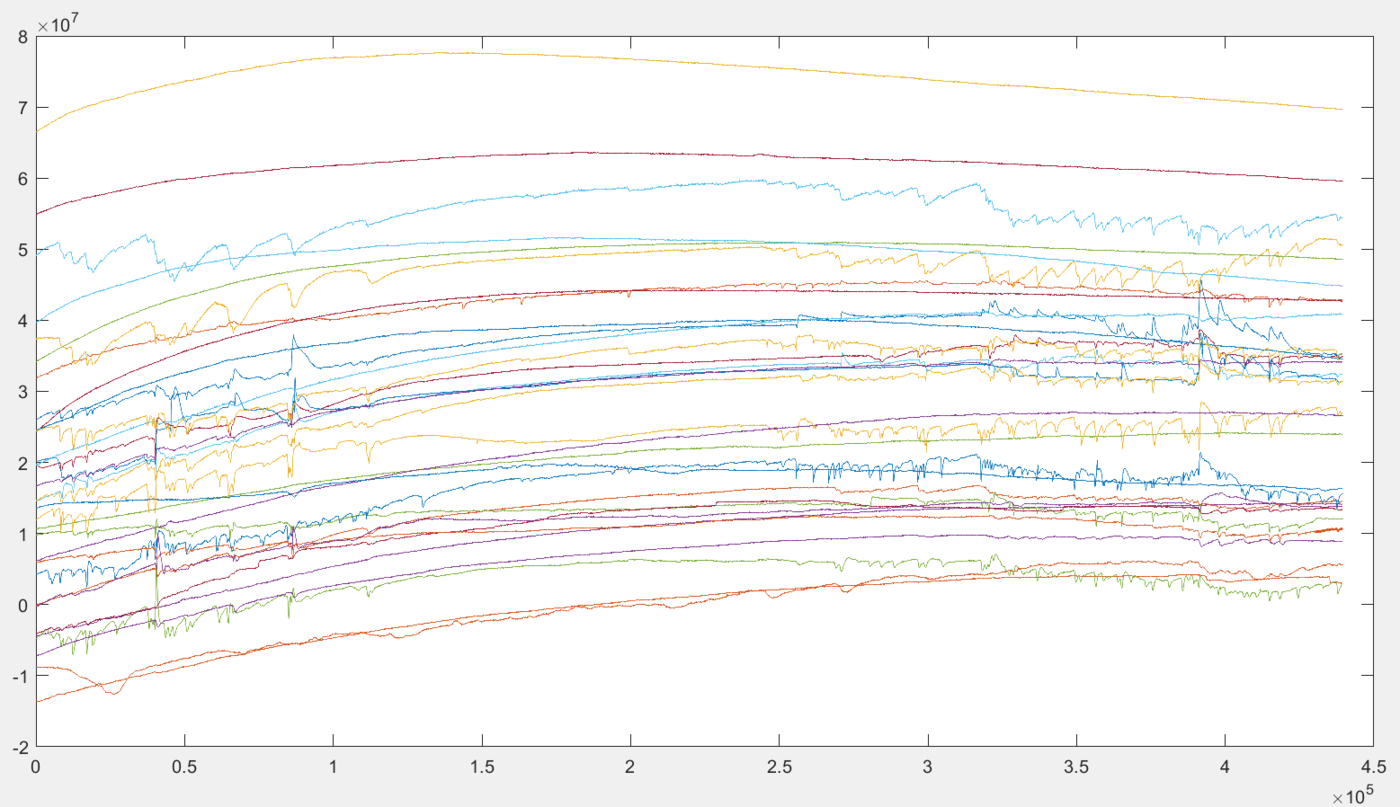}
\caption{MI tasks for Acquisition}
\label{fig_chi_dot}
\end{figure}

Finger and elbow movements are used to increase the number of activations in the sensory-motor cortex. Time intervals in which the tasks are performed were kept arrhythmic so that the amplitude of activations are preserved which otherwise tend to wane out when performing a repetitive action. 

\subsection{Subject Training}
Focused thought is critical to design a BCI system of requisite accuracy as
diffused thought process induces noise in the acquired data.
The training of the subjects in this experiment were performed with the assistance of a 
cognitive suite called the Xavier Control Panel ,which comes bundled with the Emotiv SDK.
The subjects train to focus on the movements of a virtual box under minimal sensory distractions to enable
efficient MI task execution. 

\subsection{EEG and Brain Data}
The brain activity is recorded with the assistance of a commercially available EEG headset, called EPOC+ by Emotiv Inc.(Fig. 3) which provides 14-channel EEG data.
 
\begin{figure}[h]
\centering
\includegraphics[width=3 in]{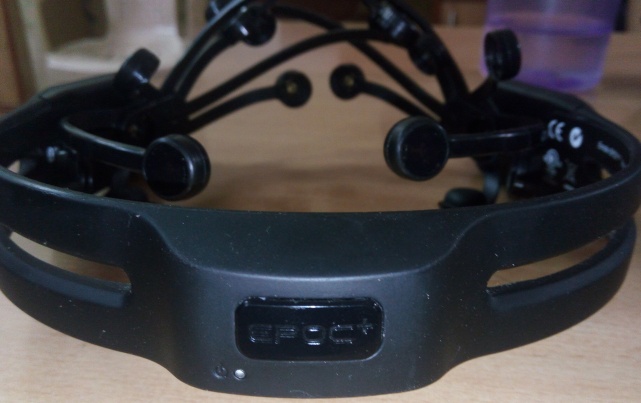}
\caption{EPOC+ Headset (14-channel)}
\label{fig_chi_dot}
\end{figure}
The EEG bands in the human brain activity are usually classified as :
\begin{itemize}

\item \textbf{Delta} - Delta band a frequency of 3 Hz or below. It tends to be the highest in amplitude and the slowest waves.

\item \textbf{Theta} - This band has a frequency of 3.5 to 7.5 Hz and is classified as "slow" activity.

\item \textbf{Alpha} - This region is localized between 7.5 and 13 Hz and is usually best seen in the posterior regions of the head on each side.

\item \textbf{Beta} - Beta activity is "fast" activity and has frequency of 14 Hz and greater.

\item \textbf{Gamma} - A gamma signal is a pattern of neural oscillation in humans with a frequency between 25 and 100 Hz, though 40 Hz is typical.

\end{itemize}
\subsection{Algorithms}
As discussed in the previous sections, in this study we have used a combination of common spatial paradigm (CSP) for feature extraction and linear discriminant analysis (LDA) for model training. The algorithm definitions and appropriate graphical analysis are discussed in this section.
\subsubsection{Commom Spatial Paradigm (CSP)}

CSP algorithm disintegrates the signal into additive components so that they have maximal variance difference in two windows.

EEG signals are usually formalized as :
\begin{equation}
\{\textbf{\textit{E}}_n\}^N \in \mathbb{R}^{\textit{ch $\times$ time}}
\end{equation}
The range of time trials varied from 100 to 300 seconds. Now, to apply EEG signals for classification, we must transform them first. The transformation to a feature vector takes place as follows :

\begin{equation}
\textbf{\textit{E}}_n\ \in \mathbb{R}^{\textit{ch $\times$ time}} \longmapsto \textit{\textbf{x}}_n \in \mathbb{R}^d
\end{equation}
Major points of concern here are :
\begin{itemize}
\item Noise reduction has to be done
\item Frequency band selection for optimal performance
\item Channel selection
\end{itemize}
Feature matrix then could be obtained as :
\begin{equation}
\textbf{\textit{X}} \in \mathbb{R}^{\textit{d$\times$ N}}
\end{equation}

The purpose of using a spatial filter like CSP, in this study, is that the signals provided by the algorithm are easier to classify even with simple methods. The objective of this section is to design spatial filters that result in optimal variances for the classification of two motor imagery signals related to left and right arm movements. The CSP filter (by Muller-Gerking \textit{et al.}) is mathematically written as :

\begin{equation}
\textit{\textbf{S}} = \textbf{\textit{W}}^\textit{T} E
\end{equation}

where \textbf{\textit{W}} $\in$ $\mathbb{R}^{\textit{d$\times$ ch}}$ is a spatial filter matrix and \textbf{\textit{S}} $\in$ $\mathbb{R}^{\textit{d$\times$ time}}$ is the filtered signal matrix. The fundamental of CSP is to maximize Eq. 5  :
\begin{equation}
 tr\textbf{\textit{W}}^T\Sigma_1\textbf{\textit{W}}
\end{equation}
with subject to Eq. 6
\begin{equation}
\textbf{\textit{W}}^T(\Sigma_1 + \Sigma_2)\textbf{\textit{W}} = \textbf{\textit{I}}
\end{equation}
where,

\begin{equation}
\Sigma_1 = Exp \quad \frac{\textbf{E}_n\textbf{E}_n^T}{tr\textbf{E}_n\textbf{E}_n^T} \quad E_n \in \{class 1\}
\end{equation}

\begin{equation}
\Sigma_2 = Exp \quad \frac{\textbf{E}_n\textbf{E}_n^T}{tr\textbf{E}_n\textbf{E}_n^T} \quad E_n \in \{class 2\}
\end{equation}
Above equations are solved with the help of \textbf{generalized eigen value problem}. Initially, we decompose as :

\begin{equation}
\Sigma_1 + \Sigma_2 = \textbf{\textit{UDU}}^\textit{T}
\end{equation}
where, \textbf{\textit{U}} is a collection of eigenvectors, and \textbf{\textit{D}} is a diagnol matrix of eigenvalues. Next we try to find the value of \textbf{\textit{P}} = $\sqrt{\textbf{D}^{-1}}\textbf{U}^\textit{T}$, and :

\begin{equation}
\hat{\Sigma}_1 = \textbf{\textit{P}}\Sigma_1 \textbf{\textit{P}}^\textit{T}
\end{equation}
\begin{equation}
\hat{\Sigma}_2 = \textbf{\textit{P}}\Sigma_2 \textbf{\textit{P}}^\textit{T}
\end{equation}

Now, any orthonomral matrices \textbf{\textit{V}}satisfy the condition

$\textbf{\textit{V}}^T$($\hat{\Sigma}_1 + \hat{\Sigma}_2$)\textbf{\textit{V}} = \textit{\textbf{I}}
Finally, we disintegrate as :
\begin{equation}
\hat{\Sigma}_1 = \textbf{\textit{V}}\Lambda \textbf{\textit{V}}^\textit{T}
\end{equation}

where, \textbf{\textit{V}} is a collection of eigenvectors, and \textbf{\textit{$\Lambda$}} is a diagnol matrix of eigenvalues. The CSP filter set is obtained as :

\begin{equation}
\textbf{\textit{W}} = \textbf{\textit{P}}^T\textbf{\textit{V}}
\end{equation}
The descriptions would be :
\begin{equation}
\textbf{W}^T\textbf{$\Sigma_1$}\textbf{W} = \textbf{$\Lambda$} = 
\begin{bmatrix}
\lambda_1 &&\\
&\ddots \\ 
&&\lambda_{ch}\\
\end{bmatrix}
\end{equation}

\begin{equation}
\textbf{W}^T\textbf{$\Sigma_2$}\textbf{W} = \textbf{\textit{I}} - \textbf{$\Lambda$} = 
\begin{bmatrix}
1 - \lambda_1 &&\\
&\ddots \\ 
&&1 - \lambda_{ch}\\
\end{bmatrix}
\end{equation}

where $\lambda_1$ $\geq$ $\lambda_2$ $\geq$ $\cdots$ $\geq$ $\lambda_{ch}$. Hence, the first CSP filter $\omega_1$ provides maximal variance for $1^{st}$ class, and the last filter $\omega_2$ for $2^{nd}$ class. The first and last \textit{m} filters are selected in the following manner :

\begin{equation}
\textbf{\textit{W}}_{csp} = (\omega_1 \quad \cdots \quad \omega_m \quad \omega_{ch-m+1} \quad  \cdots \quad\omega_{ch}) \in \mathbb{R}^{2m \times ch}
\end{equation}

and the filtered signal mathematically is :
\begin{equation}
\textbf{\textit{s}}\textit{(t)} = \textbf{\textit{W}}_{csp}^\textit{T}\textbf{e}\textit{(t)} = (s_1(t) \quad \cdots \quad s_d(t))^T,
\end{equation}

i.e., \textit{d = 2m}.

Feature vector \textit{x} = ($x_1$, $x_1$, \dots, $x_d$)$^T$, is then calculated as :

\begin{equation}
\textbf{x}_i = \log\bigg(\frac{var[s_i(t)]}{\sum_{i=1}^{d}var[s_i(t)]}\bigg)
\end{equation}

Now that we have our feature space constructed, we wo ahead by beginning the model training using LDA.

\subsection{Linear Discriminant Analysis}
Fisher's linear discriminat analysis (LDA), a very popular binary classifier, is based on mean vectors and covariance matrices of patterns of each individual class. Here, we attempt to convert a \textit{d}-dimensional vector \textbf{$x$} to a scalar \textit{z} as :

\begin{equation}
\textbf{z} = \textbf{w}^\textit{T}\textbf{x}
\end{equation}

Basically, the LDA provides us with an optimal projection $w$ so that $z$ becomes easy to discriminate. The fundamental of LDA (criterion) is maximizing :

\begin{equation}
\textbf{J}(w) = \frac{(m_1 - m_2)^2}{s_1 + s_2},
\end{equation}

where,
\begin{itemize}
\item $m_1$ and $m_2$ are averages for $z_n$ $\in$ {class 1} and $z_n$ $\in$ {class 2} respectively

\item $s_1$ and $s_2$ are variances for $z_n$ $\in$ {class 1} and $z_n$ $\in$ {class 2} respectively

\end{itemize}

The variables are so defined as :

\begin{equation}
(m_1 - m_2)^2 = (\textbf{w}^T\mu_1 - \textbf{w}^T\mu_2)(\textbf{w}^T\mu_1 - \textbf{w}^T\mu_2)^T
\end{equation}

\begin{equation}
(s_1 + s_2) = \textbf{w}^T\Sigma_1\textbf{w} + \textbf{w}^T\Sigma_2\textbf{w}
\end{equation}

where,
\begin{itemize}
\item $\mu_1$ and $\mu_2$ are averages for $x_n$ $\in$ {class 1} and $x_n$ $\in$ {class 2} respectively

\item $\Sigma_1$ and $\Sigma_2$ are variances for $x_n$ $\in$ {class 1} and $x_n$ $\in$ {class 2} respectively
\end{itemize}
The cost function $J(w)$, can then be written as :

\begin{equation}
\textbf{J}(w) = \frac{\textbf{w}^T\textbf{S}_B \textbf{w}}{\textbf{w}^T\textbf{S}_W \textbf{w}}
\end{equation}
where,

\begin{equation}
\textbf{S}_B = (\mu_1 - \mu_2)(\mu_1 - \mu_2)^T
\end{equation}

\begin{equation}
\textbf{S}_W = \Sigma_1 + \Sigma_2
\end{equation}
and then the final solution is given by the following :

\begin{equation}
\hat{\textbf{w}} \propto \textbf{S}_W^{-1}(\mu_1 - \mu_2)
\end{equation}
Finally, an appropriate $z_0$ threshold is chose for accurate categorization of any $x$ by :

\begin{equation}
\hat{\textbf{w}}^T x \geq z_0 \to \textbf{x} \in \{class 1\}
\end{equation}
\begin{equation}
\hat{\textbf{w}}^T x < z_0 \to \textbf{x} \in \{class 2\}
\end{equation}
A classification example from our testing is shown below in Fig. 4 :

\begin{figure}[h]
\centering
\includegraphics[width=3.3 in]{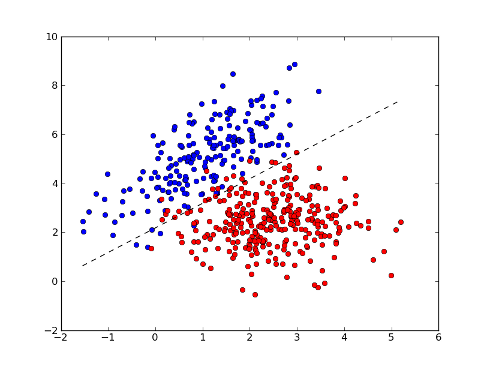}
\caption{Binary Classification using LDA}
\label{fig_chi_dot}
\end{figure}

The overall workflow of the process is demonstrated by the flow diagram in Fig. 5 :

\begin{figure}[h]
\centering
\includegraphics[width=3.3 in]{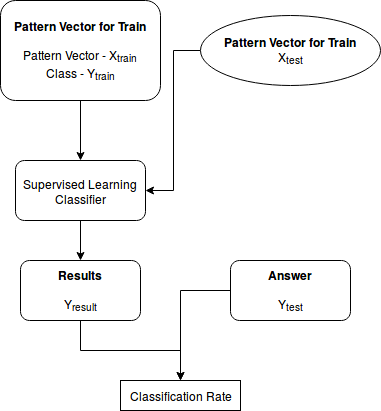}
\caption{Workflow for EEG Classification}
\label{fig_chi_dot}
\end{figure}

\subsection{Support Vector Machine with Non-Linear Kernel}

With the LDA-based approach we were abale to achieve a classification rate of 89\% for a 2-class input system. However, for a multi-class categorization, we cannot use LDA, so we used non-linear SVM classification technique and implemented it for 4-class inputs. SVM was used primarily because it has avery small memory footprint and is almost real-time when put to test.

Suppose we have a dataset {$X^{(i)}$, $Y^{(i)}$}, i = 1,2,$\dots$,m and X $\in$ $\mathbb{R}^d$, Y $\in$ $\mathbb{R}^1$ with the seperating hyperplane defined as $W^T X + b = 0$ such that, 
\begin{equation}
\textbf{W}^T \textbf{X} + b > 0 \quad if \quad Y^{(i)} = +1
\end{equation}
\begin{equation}
\textbf{W}^T \textbf{X} + b < 0 \quad if \quad Y^{(i)} = -1
\end{equation}

Now we suppose that the training data (linearly seperable) satisfy the following conditions :
\begin{equation}
\textbf{W}^T \textbf{X} + b \geq +1 \quad for \quad Y^{(i)} = +1
\end{equation}
\begin{equation}
\textbf{W}^T \textbf{X} + b \leq -1 \quad for \quad Y^{(i)} = -1
\end{equation}

Combining the above to a single condition, we get :
\begin{equation}
Y^{(i)}(\textbf{W}^T X + b) \geq 1 \quad for \quad \forall i
\end{equation}

Our task now is to find a hyperplane(W,b) with maximal distance between itself and the closest data points, while obeying the mentioned constraints. Mathematically :
\begin{equation}
max (\frac{2}{||\textbf{W}||}) w.r.t. (W,b)
\end{equation}

The trick to designing a SVM is to solve the DUAL of the above inequality. It can be proved that :
\begin{equation}
min \quad \textit{primal} = max(minL(W,b,\alpha))
\end{equation}
where the primal problem is defined as :

\begin{equation}
min \bigg(\frac{\textbf{W}^T\textbf{W}}{2}\bigg)
\end{equation}

To begin solving our above problem, we construct the Langrangian : 

\begin{equation}
L(W,b,\alpha) = \frac{1}{2}||\textbf{W}||^2 - \sum_{i=1}^{m}\alpha_i[Y^{(i)}(\textbf{W}^T X^{(i)} + b) - 1]
\end{equation}
where, $\alpha$ is a Langragian multiplier with the condition $\alpha_i \geq 0$. We have to minimize it w.r.t. W and b;we set the respective derivatives equal to zero. Derivative w.r.t. W and b if set to zero:
\begin{equation}
\textbf{W} = \sum_{i=1}^{m}\alpha_iY^{(i)}X^{(i)}
\end{equation} 
\begin{equation}
\sum_{i=1}^{m}\alpha_iY^{(i)} = 0
\end{equation}
Putting the results back into the Lagrangian leaves us with :
\begin{equation}
\textbf{L}(W,b,\alpha) = D(\alpha) = \sum_{i=1}^{m}\alpha_i - \frac{1}{2}\sum_{i,j=1}^{m}Y^{(i)}Y^{(j)}\alpha_i\alpha_j(X^{(i)})^T(X^{(j)})
\end{equation}
Now the DUAL problem reduces to the eq.:
\begin{equation}
max \quad \textbf{D}(\alpha) = \sum_{i=1}^{m}\alpha_i - \frac{1}{2}\sum_{i,j=1}^{m}Y^{(i)}Y^{(j)}\alpha_i\alpha_j(X^{(i)}X^{(j)})
\end{equation}
Solving the above optimization would give us $\alpha_i$. Moreover, the Karush-Kuhn-Tucker condition is satisfied on this solution :

\begin{equation}
\alpha_i[Y^{(i)}(W^TX^{(i)})-1] = 0 \quad for \quad i=1,2,\dots,m
\end{equation}
Now, W and b could be found using respectively :

\begin{equation}
\textbf{W} = \sum_{i=1}^{m}\alpha_iY^{(i)}X^{(i)}
\end{equation}
\begin{equation}
\textbf{b} = -\frac{max_{i:Y^{(i)}=-1}W*^TX^{(i)} + min_{i:Y^{(i)}=1}W*^TX^{(i)}}{2}
\end{equation}
Now if the input feature space is non-linearly seperable, we map the input feature space into another space where it can be classified linearly. We assume that $\Phi : X \to F$ be a non-linear map form input X to a higher dimensional feature space, F. So, the inner product $\langle X^{(i)},X^{(j)} \rangle$ in the higher dimensions is $\langle \phi^{(i)},\phi^{(j)} \rangle$.

The easiest way to compute the inner product in the feature space (higher dimensional space) is by using the Kernel Function. It is defined as :

\begin{equation}
\textbf{K}(x,z) = \langle \phi^{(i)},\phi^{(j)} \rangle
\end{equation}
Graphically, the transformationm would look something as shown in Fig. 6.

\begin{figure}[h]
\centering
\includegraphics[width=3.4 in]{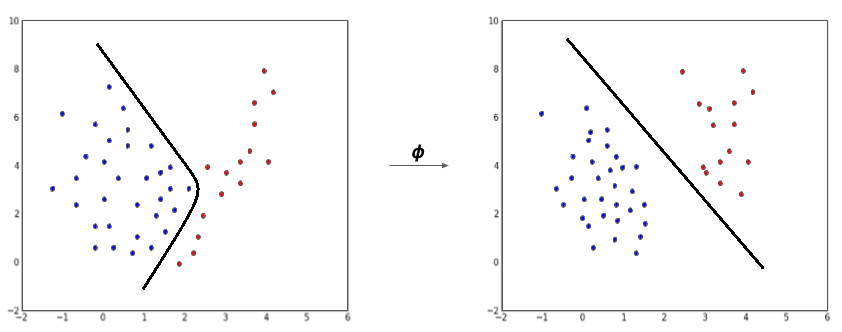}
\caption{Non-linearly seperable data (on the left) projected onto a space where it is linearly seperable (on the right) using a non-linear Kernel Function.}
\label{fig_chi_dot}
\end{figure}

So, the optimization and decision functions are rendered respectively as :
\begin{equation}
max \quad \textbf{D}(\alpha) = \sum_{i=1}^{m}\alpha_i - \frac{1}{2}\sum_{i,j=1}^{m}Y^{(i)}Y^{(j)}\alpha_i\alpha_jK\langle X^{(i)}X^{(j)}\rangle
\end{equation}
\begin{equation}
\textbf{F}(X) = Sign(\sum_{i=1}^{m}\alpha_iY^{(i)}K(X^{(i)},X)+b)
\end{equation}
The above equations are solved with appropriate choice of the kernel function, which in our case, we chose a Polynomial Kernel Function of degree \textit{d}, defined as:
\begin{equation}
\textbf{K}(X,Y) = (X^TY + 1)^d
\end{equation}
Solving the above equations, we get the classification/prediction label for each time stamped EEG data.
\section{HARDWARE INTERFACE}
In order to demonstrate the real-time computational capabilities of our BCI, we integrated it with a fixed wing UAV (2-command control) and to a multi-rotor (4-command control).
\subsection{Delta Wing}

The testing was done on a prototype first, before deploying it in actual flight. 
The output from the prediction framework (LDA based 2 class classification) is sent to a microcontroller unit. An \textit{Arduino} (ATmega328P) board
is used to control a prototype \textit{elevon} of a delta configuration made of chloroplast. The model is pre-programmed to do a particular action with the help of servo motors depending upon the
type of input commands received. The fig below (Fig. 7) shows the working model.
\begin{figure}[h]
\centering
\includegraphics[width=3.3 in]{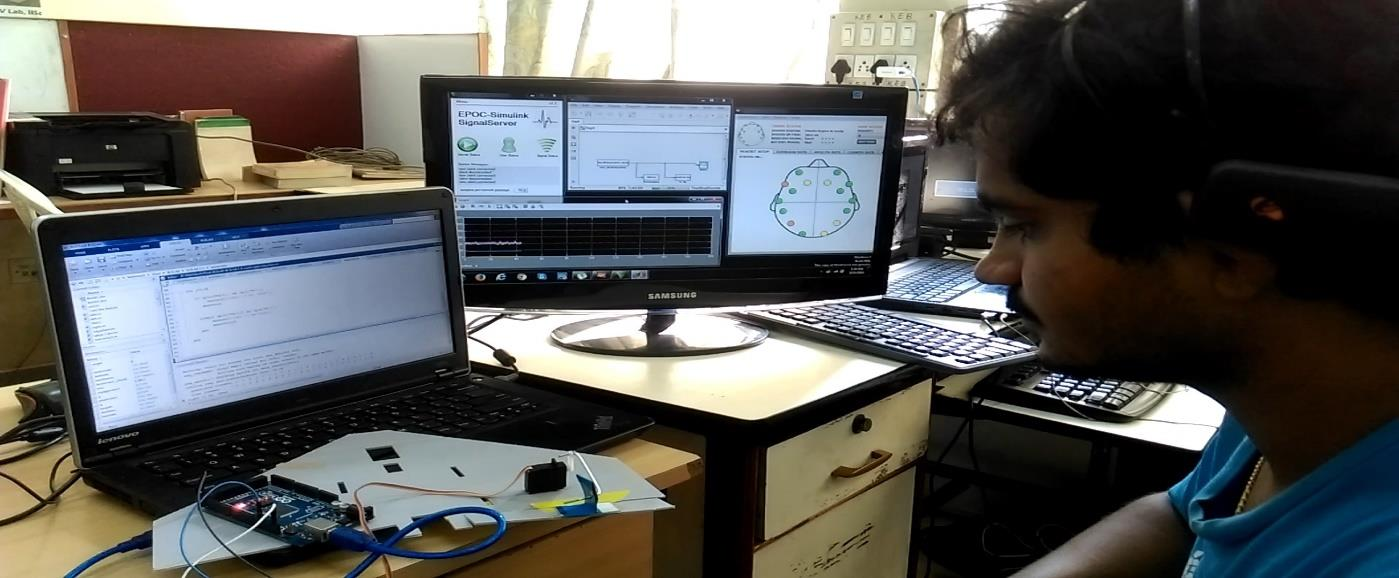}
\caption{Prototype Demonstration}
\label{fig_chi_dot}
\end{figure}

\subsection{Multi-Rotor}
In order to establish a proof of concept , an off-the-shelf quadrotor UAV platform (Fig. 8) compatible with the Python
programming language was used. The algorithm, also written in Python, is hosted on a ground station that is in constant duplex communication with the UAV via WiFi. The UAV host an array of flight instruments such as
\begin{itemize}
\item front camera
\item bottom camera (low field-of-view)
\item 3-axis accelerometer
\item 3-axis gyroscope
\item a magnetometer
\item barometer
\item ultrasonic sensor
\item motherboard
\end{itemize}

\begin{figure}[h]
\centering
\includegraphics[width=2.9 in]{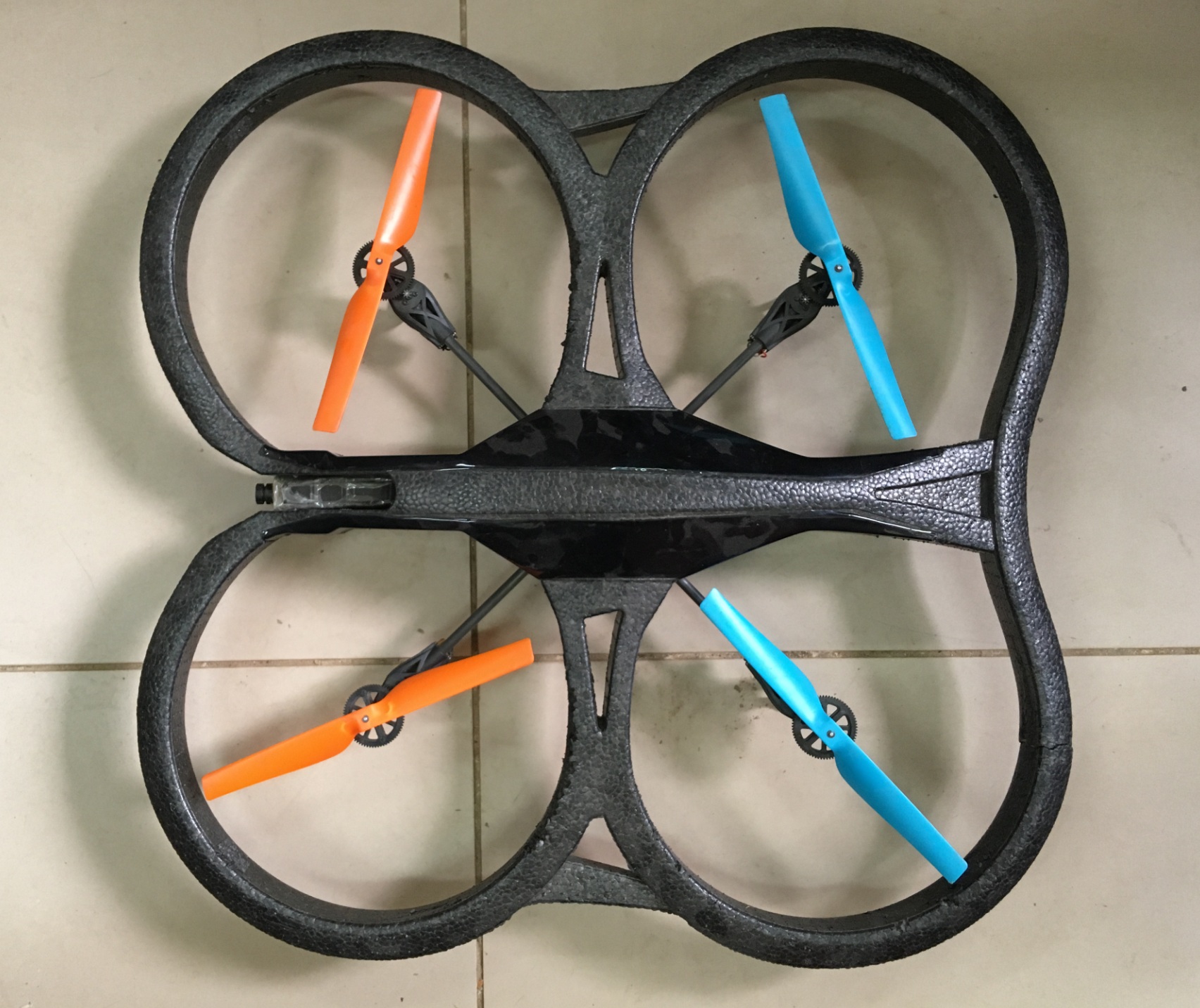}
\caption{AR Parrot 2.0 used for testing}
\label{fig_chi_dot}
\end{figure}

The inner control loop program is embedded onto the motherboard while the outer navigation loop is dictated by an open-source Python library called python-ardrone which is hosted on the ground station. Sensor data fusion is achieved by using Extended Kalman Filtering (EKF) method extensively. The implemented algorithm sends only high-level commands (NLSVM based 4-class classification) to the UAV platform (Fig. 9) so that the inherent stability is not compromised.

\begin{figure}[h]
\centering
\includegraphics[width=2.9 in]{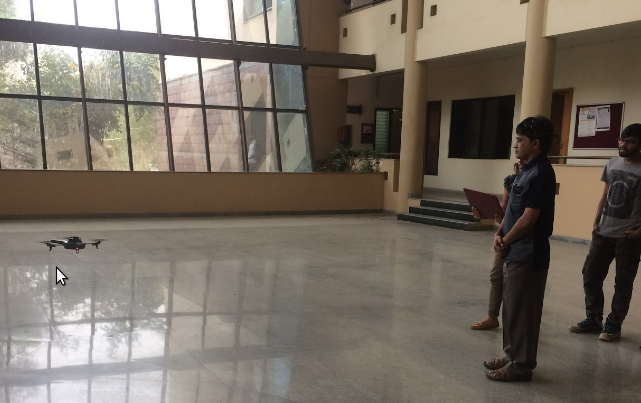}
\caption{Real-time testing}
\label{fig_chi_dot}
\end{figure}

\section{RESULTS}
The interface works pretty well in real time [18], at a rate of 90 Hz.
Individual results of each subject have been tabulated below :

\captionof{table}{Performance Evaluation} \label{tab:title} 
\centering
\begin{tabular}{ccccc}
  \hline
  Subject &  A$_{LDA}$ & A$_{NLSVM}$ &T$_{focus}$&T$_{max}$\\
  \hline
1 & 81\% & 79\% & 150 s & 287 s\\
2 & 83\% & 83\% & 143 s & 295 s\\
3 & 82\% & 79\% & 156 s & 304 s\\
4 & 89\% & 83\% & 151 s & 309 s\\
5 & 85\% & 80\% & 144 s & 322 s\\
6 & 77\% & 78\% & 139 s & 254 s\\
7 & 82\% & 81\% & 149 s & 322 s\\
8 & 79\% & 77\% & 155 s & 330 s\\
9 & 87\% & 82\% & 129 s & 310 s\\
10* & 98\% & 96\% & 133 s & 505 s\\
11 & 91\% & 88\% & 143 s & 303 s\\
12 & 92\% & 90\% & 166 s & 367 s\\
13 & 90\% & 85\% & 162 s & 332 s\\
  
  \hline
  \end{tabular}
\justify
where, A$_{LDA}$ represents the testing accuracy for LDA (2-class), A$_{NLSVM}$ represents the testing accuracy for Non-Linear SVM (4-class) classifier, T$_{focus}$ represents the average time taken to focus on a certain motor imagery task and T$_{max}$ is the maximum focus time. 

*The increased performance in case of subject 10 could be attributed to the fact that the subject had been doing \textit{Yoga} for some years. Yoga has been known to enhance mental and physical performance in terms of memory, focus duration (Akhtar \textit{et al.} [16]) and physical stabililty (Omkar \textit{et al.} [17]).

\section{CONCLUSIONS}
\justify
A system was developed which takes EEG (electroencephalography) signals as input, modifies
the signal for feature extraction and interfaced with \textit{elevon} for controlling it wireless connection. The
raw EEG data was extracted from the brain of the subject using the EMOTIV EPOC+ headset. The raw EEG data is a result of only intuitive thinking without any actual
physical movements. The data was then transformed in a suitable data form to be processed. Further removal of artefacts, unwanted frequencies and irregular data was done. The processed data was then used for
preparing the model for machine learning using LDA analysis method. Suitable markers to
mark the Event Related Potential (ERP) to train the machine for evaluate were added. A twin
dataset is applied to the created model, to calculate the misclassification and the
error percentage in the same dataset and the twin dataset respectively. The error percentage is
11\%. An offline analysis was done by encoding the signals into commands for the fixed wing
Elevon. The final interfacing was done with the Elevon and the repeated creation, testing,
evaluation, and deployment of the models were done to reach this accuracy. The BCI is also tested on an off-the-shelf multi-rotor, with classification accuracy as high as 90\%, for 4-class based control. This work can
further be extended to control other kinds of UAV and the complexity can be increased and customized based on the requirement. The modularity, remote access and control of interfaced
Elevon based on pure brain signals in a BCI system is demonstrated.

\addtolength{\textheight}{-12cm}   

\section*{ACKNOWLEDGEMENTS}

We give warm thanks to \textit{Mehvesh Ibrahim}, \textit{Satya Shree}
and \textit{Kapil Bharadwaj} for providing assistance with the data collection for the Delta-Wing. We also thank \textit{Priya Rao}, \textit{Ankita Verma}, \textit{Akshay Khokkar} and \textit{Likith Reddy} from NIT Sringar, for helping collect data and testing the classification on the multi-rotor. We also extend our regards to \textit{Navaneethkrishnan B} for the helpful discussions.

\end{document}